\documentclass{article}
\usepackage{ijcai22}

\usepackage{times}
\usepackage{soul}
\usepackage{url}
\usepackage[hidelinks]{hyperref}
\usepackage[utf8]{inputenc}
\usepackage[small]{caption}
\usepackage{graphicx}
\usepackage{amsmath}
\usepackage{amsthm}
\usepackage{booktabs}
\usepackage{algorithm}
\usepackage{algorithmic}
\usepackage{mathrsfs}
\usepackage{amsfonts,amssymb}

\usepackage{graphicx}
\usepackage{diagbox}
\usepackage{makecell}
\usepackage{multirow} 
\usepackage{tikz}
\usepackage{comment}
\usepackage{amsmath,amssymb} % define this before the line numbering.
\usepackage{color}
\usepackage{subfigure}

\usepackage[accsupp]{axessibility}  % Improves PDF readability for those with disabilities.

\title{Unrestricted Adversarial Samples Based on Non-semantic Feature Clusters Substitution}

\author{
MingWei Zhou$^1$
\and
Xiaobing Pei$^2$\and
Haoxi Zhan$^3$\And
Fourth Author$^4$
\affiliations
$^1$Huazhong University of Science and Technology\\
$^2$Second Affiliation\\
$^3$Third Affiliation\\
$^4$Fourth Affiliation
\emails
\{first, second\}@example.com,
third@other.example.com,
fourth@example.com
}
\begin{document}

\maketitle

\begin{abstract}
Most current methods generate adversarial examples with the $L_p$ norm specification. As a result, many defense methods utilize this property to eliminate the impact of such attacking algorithms. In this paper,we instead introduce “unrestricted” perturbations that create adversarial samples by using spurious relations which were learned by model training. Specifically, we find feature clusters in non-semantic features that are strongly correlated with model judgment results, and treat them as spurious relations learned by the model. Then we create adversarial samples by using them to replace the corresponding feature clusters in the target image. Experimental evaluations show that in both black-box and white-box situations. Our adversarial examples do not change the semantics of images, while still being effective at fooling an adversarially trained DNN image classifier.
\end{abstract}

\section{Introduction}

Deep learning has been widely used today, such as face recognition, intelligent robots, malware detection and so on. The continuous growth of computer computing resources and the explosive growth of data volume make deep learning unique in terms of feature learning. Although deep learning has achieved great success in many fields, recent studies have shown that even minor perturbations can fool well-trained convolutional neural networks with strong generalization ability.\cite{szegedy2013intriguing,goodfellow2014explaining}. This raises concerns about the security of deep learning applications. This requires an in-depth study of adversarial examples to help better understand the potential vulnerabilities of ML models and improve their robustness. To date, various approaches have been proposed to generate adversarial examples\cite{goodfellow2014explaining,carlini2017towards,kurakin2018adversarial,xiao2018generating} ,and many of these attacks search for perturbations in a finite  $L_P$ norm to keep their photos authentic. However, it is well known that $L_p$  norm distance as a measure of perceptual similarity is not ideal \cite{johnson2016perceptual,isola2017image}.In addition, recent studies have shown that under new unknown attacks, defense methods based on $L_p$ norm perturbation training are completely not robust\cite{kang2019testing} . Therefore, exploring different adversarial examples, especially those with "unrestricted" perturbation amplitude, has received considerable attention in academia and industry \cite{brown2018unrestricted}. Recently, the work based on generative adversarial networks (GANs) \cite{goodfellow2014explaining} introduced unrestricted attacks\cite{xiao2018generating}. A number of approaches have been developed to mitigate the threat of adversarial examples. These include increasing training data with adversarial examples\cite{goodfellow2014explaining,kurakin2016adversarial,madry2017towards,sinha2017certifiable}, removing adversarial perturbation\cite{gu2014towards,song2017pixeldefend,samangouei2018defense}, and encouraging the smoothness of classifiers\cite{cisse2017parseval}. Recently, \cite{raghunathan2018certified,wong2018provable} proposed defense methods based on the theoretical proof of minimizing the upper bound of training loss under the worst-case perturbation. Although inspired by different viewpoints, a common design principle of current defense methods is to make the classifier more robust to the small perturbation of its input. There are also defenses against "unrestricted" perturbation, such as defense models in unrestricted confrontations\cite{brown2018unrestricted}. The robust accuracy of the best LLR and TRADESv2 against perturbation, space attack and SPSA attack is close to 100 \% , but the robust accuracy of LLR and TRADESv2 rapidly drops to about zero after CAA writes these attacks. The results show that the existing unrestricted adversarial defense model seriously overfits a single test attacker \cite{mao2021composite}.

In this paper, we use an attacking strategy that is not restricted by $L_P$  norm. The image style can be different and the size can be changed, but as long as it can be recognized as the same category by humans,  we regard this adversarial sample as legitimate. At the same time, the previous methods of generating adversarial samples are to add perturbations on the image, and the essence of adding perturbations is to change the features of the image. Therefore, this paper directly starts from the feature level to generate adversarial samples based on feature substitution, rather than finding the perturbations corresponding to these features one by one. However, it is difficult to generate adversarial samples based on feature substitution. Previous algorithms usually choose to add small noise.As long as the perturbations were small enough, the semantics would not change in the view of human beings. How do we ensure that the image features are replaced and the semantic information of the image remains unchanged? In this paper, we use the latest developments in style transfer models \cite{goodfellow2014generative,odena2017conditional,gulrajani2017improved} and the models trained in papers that argue that adversarial perturbations are not bugs but features. Specifically, we find the appropriate target class style features from the robustness feature dimension and the non-robust feature dimension, and then replace the target class style features on the attacked image. Since only the style feature is changed without changing the content feature, and the semantics of the image is not changed, human beings can normally identify the image category. Meanwhile, as the parameter structure of the model is not required  when attacking the model,  it belongs to the black-box attack method. We also tested attacks on several state-of-the-art defenses, not to prove that our attacks can break these defenses, but to prove the novelty of the  feature-based replacement. Our attack based on feature substitution provides another perspective on the vulnerability of deep learning models, encouraging the model to improve the robustness of the model from the feature level.

In conclusion, our contributions are as follows :

1)We propose a new attack pattern based on feature replacement, and consider it from the aspects of robustness and non-robustness.

2)We propose a new attack method based on feature substitution of style transfer, which can well preserve the semantics of images.

3)We use this method to carry out both targeted and non-targeted black-box attacks on the CIFAR10 dataset \cite{krizhevsky2009learning} and conduct experiments to show their effectiveness.

4)We attack several state-of-the-art defense technologies and prove that our proposed method has better transferability  and is difficult to defend.

\section{Related Works}
According to the topics of this paper, we briefly review previous works on CNN ' s adversarial attack and defense methods, as well as image style transfer models.

\subsection{Adversarial Attacks and Defense for CNN}

While many adversarial attacks are based on  $L_p$ norm, many attack methods that are not restricted by $L_p$ norm have been proposed recently. For example, the method of obtaining adversarial samples through spatial transformation \cite{xiao2018spatially}, the method of directly generating adversarial samples through GAN structure \cite{song2018constructing}, the method of generating adversarial samples by semantic editing by CGAN structure \cite{joshi2019semantic}, and the method of applying perturbations in feature map interpolation, which has lower additional reduction in image quality under the reasonable limitation of interpolation coefficient \cite{qiu2020semanticadv}. This paper is based on the fact that it is not restricted by $L_p$ norm.

While white-box attacks have achieved remarkable performance,  black-box attacks are still ineffective or only under certain restrictions. For example, single pixel attacks\cite{su2019one} and expected transformation attacks\cite{athalye2018synthesizing}  do not work well on complex images. The UPSET and ANGRI method  only works well when it knows what kind of network structure the model uses. When it needs to transfer, the effect is not good. ZOO\cite{chen2017zoo} is used only when you know the confidence of the model input image decision. The method used in this paper is also black-box, but still works well on complex images without knowing what kind of network structure the attacked model is and how much confidence the model has in the input image.

The current adversarial attacks are based on perturbations, but the perturbation-based attacks need to find the corresponding perturbations of the feature one by one. In this paper, we start directly from the features, positioning features to replace. Madry et al. \cite{ilyas2019adversarial} put forward a very important point that adversarial samples are non-robust features instead of bugs, which separate the robust features from the non-robust features. This paper is based on the robust feature dimension and non-robust feature dimension to find the replaceable feature group, so as to replace the features directly.

The existing defense methods usually attempt to make the classifier more robust to the small perturbations of the image. There is an ‘arms race’ between increasingly complex attack and defense methods. As indicated in \cite{athalye2018obfuscated}, the strongest defense so far has been adversarial training\cite{mkadry2017towards}. In this paper, we focus on two different adversarial training defense models, namely PGD adversarial training defense model\cite{carlini2017towards} and IAT adversarial training defense model\cite{lamb2019interpolated} .

Existing strategies for applying style transfer to adversarial samples include strategies for using style transfer to resist camouflage\cite{duan2020adversarial}  and strategies for transferring textures to attack\cite{bhattad2019unrestricted}. Among them, the strategy of using style transfer to resist camouflage is to generate the adversarial perturbations of the physical world. Under the principle of not changing semantics, the pistol can be forged as toilet paper, and the traffic sign can be forged as a barbershop. However, only some images can be targeted attacked with this style transfer method. And there is no way to realize the transferability of style transfer attacks. The strategy of transferring texture to attack is based on small perturbations, which is a white-box attack method. As the size of perturbations is limited, the effect is also limited, and more complex calculations are needed, so the efficiency is not satisfactory.
\subsection{Style Transfer Model}
Image style transfer was first used for rendering images\cite{kyprianidis2012taxonomy}, the main principle is to synthesize and transform textures\cite{elad2017style}. Gatys et al. realized the different types of features learned by CNN in deep and shallow layers, so they used this to separate the content and style of the image. Then they are recombined to produce artistic images such as waves and shout styles\cite{gatys2016image}. Since the content and style of the image are separated, changing the style of the image can still retain the semantics of the image. Our goal is to replace the style of the image with the style of another image to generate an adversarial sample without changing the semantics, nor explicitly restricting the $L_P$ norm of the perturbations. In order to generate our style transfer adversarial samples, we use a pre-trained VGG19 network  to extract style features. We directly replace the style features of the attack image with the style features of the target image. The method of replacing the style features is to minimize correlation statistics of features in the style layer for generated images and target class images, which is to minimize the style loss.

\section{Method}
In this section, we first summarize our understanding of adversarial attacks at the feature level. Then we introduce our feature replacement method and the selection method of style features, and how the target image is generated.

\subsection{Analysis of Adversarial Attack from Feature Level}
As Madry et al\cite{carlini2017towards}point out, we can realize that adding small adversarial perturbation essentially changes the non-robust characteristics of the image. Therefore, we believe that the essence of adversarial samples is to replace the features. For instance, the previous $L_P$ norm perturbation is to replace the non-robust features with other types of non-robust features which were added with perturbation, and non-$L_P$ norm perturbation is to replace the robust features, non-robust features with the corresponding features which were added with perturbation.

This paper analyzes the recognition ability of each model from the level of robust features and non-robust features. We use the method proposed by Madry et al. \cite{carlini2017towards} to train a model that only recognizes robust features and another model for non-robust features. We use perturbation to find images that make the above two models discriminate inconsistently, the image is then handed over to the basic model, and the PGD adversarial training model, and the IAT adversarial training model for identification. Through a series of experiments, as shown in Table 3-1 and Table 3-2, we find that the robustness of the basic model, the PGD adversarial training model and the IAT adversarial training model is positively correlated with the discriminative proportion of robust features in the model, while generalization is positively correlated with the discriminative proportion of non-robust features in the model. Therefore, attacks on robust models should mainly start from the perspective of robust features. However, traditional perturbation-based attack methods often start from the perspective of non-robust features. In this paper, we directly find suitable and strong enough style features from the two dimensions of robust features and non-robust features, which include relatively key robust features, so we can better attack the robust model.
\renewcommand\tabcolsep{5.0pt} % 
\renewcommand {\thetable} {3-1}
\begin{table*}
	\begin{center}
		\caption{The proportion of categories judged by one thousand pictures}
		\begin{tabular}{llllll}
			\hline\noalign{\smallskip}
			Model & R & NR & Basic & IAT & PGDAT\\
			\noalign{\smallskip}
			\hline
			\noalign{\smallskip}
			Airplane & 1 & 998 & 998 & 565 & 620\\
			
			Ship & 999 & 0 & 1 & 435 & 373\\
			\hline
		\end{tabular}
	\end{center}
\end{table*}

In Table 3-1, R represents a model that can only recognize robust features, NR represents a model that can only recognize non-robust features, and Basic is the ResNet18 model trained on the CIFAR10 dataset, IAT and PGDAT are two adversarial training models respectively. It can be seen that the perturbed thousand images are identified as the ship category at the robust feature level and as the airplane category at the non-robust feature level.We can see the proportion of airplane categories in the identification results of the Basic model, IAT model, and PGDAT model increased in turn, so the experimental results showed that the proportion of robust features in the Basic model, IAT model, and PGDAT model increased in turn.

\renewcommand\tabcolsep{5.0pt} % 
\renewcommand {\thetable} {3-2}
\begin{table*}
	\begin{center}
		\caption{The proportion of categories judged by one thousand pictures}
		\begin{tabular}{lllllll}
			\hline\noalign{\smallskip}
			& Model & R & NR & Basic & IAT & PGDAT\\
			\noalign{\smallskip}
			\hline
			\noalign{\smallskip}
			& Accuracy & 78.69 & 82.00 & 95.28 & 91.86 & 83.53\\
			\hline
			Attack Success&pgd&62.04 & 99.90 & 98.98 & 57.24 & 53.93 \\
			\hline
		\end{tabular}
	\end{center}
\end{table*}

The meaning of each model in Table 3-2 is the same as 3-1. It can be seen that the generalization of the Basic model, the IAT model and the PGDAT model for basic samples decreases in turn, but the robustness of the Basic model, the IAT model and the PGDAT model for pgd adversarial samples increases in turn. Combined with the conclusion in Table 3-1 that the proportion of robust features in the Basic model, IAT model, and PGDAT model increases in turn, it is believed that the proportion of model robustness features is proportional to model robustness.

At the same time, we find that some robust features and non-robust features learned by the model form structures and clusters, but these feature clusters have nothing to do with semantic discrimination For example, a feature cluster that recognizes the deer category is the forest background. As long as the forest background is recognized, the classifier will determine the image category as deer with a high probability (as shown in Figure 3-1 and Figure 3-2). We can implement adversarial attacks by virtue of such substitution of feature clusters that are not related to semantic judgment. The usual perturbation-based adversarial attack methods only replace the most critical non-robust features and robust features in the image, and it is difficult to expose the identification loopholes caused by such feature clusters. 
\begin{figure}[htbp]
	\centering
	\renewcommand {\thefigure} {3-1}
	\begin{minipage}[t]{0.48\textwidth}
		\centering
		\includegraphics[width=2cm]{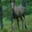}
		\caption{Deer sample with forest background}
	\end{minipage}
	\renewcommand {\thefigure} {3-2}
	\begin{minipage}[t]{0.48\textwidth}
		\centering
		\includegraphics[width=2cm]{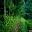}
		\caption{Pure forest background image(Both pictures 3-1 and 3-2 are judged to be deer)}
	\end{minipage}
	
\end{figure}

\subsection{Style Based Black-box Attack}
\textbf{Style selection.} To create style transfer black-box adversarial attack samples, we need to find images to extract style, which we call "style sources" (Ls). A naive strategy is to randomly select an image as the style source (Ls). Although we can also find a suitable style source (Ls) through this strategy. For example, we regard zebra stripes as the style source (Ls), which can make the image be judged as a frog category with a high probability. But it is tantamount to finding a needle in a haystack, and it is difficult to find a suitable targeted  style source (Ls) for the category. Alternatively, we can randomly select Ls from the target class we want to attack. Compared to randomly selecting Ls, this strategy is easier to find suitable Ls because we extract the style from the known target class. The style extracted by this method is more likely to contain the key style features of the target class. A better strategy for selecting Ls is to first extract the style features of the target image. Then judge the confidence through the robust feature dimension and the non-robust feature dimension, and use a certain weight to  find the Ls whose style features are judged as the target class with high confidence in both dimensions. This strategy is reasonable. On the one hand, the style features of the target class may have inconsistent categories determined by the robust feature dimension and the non-robust feature dimension. On the other hand, the confidence of the style feature to be judged as the target category may not be enough, so the influence weight on the model judgment is very small and the replacement of such style features has no effect on the model judgment. Our attack generates more aggressive images due to the selection of style features with high influence weights that do not conflict in the dimensions of robust and non-robust features.

\textbf{Style Transfer.} For style transfer, we employ cross-layer statistics. We use R11, R12, R21, R22, and R31 from the preprocessed VGG19 as the style layer and R22 as the content layer. When performing style transfer, we limit the content loss within a range, and stop when the content loss exceeds this range or the style loss no longer decreases.

Black-box adversarial methods based on style transfer directly attack images without modifying network parameters. Our overall objective function for style transfer attack consists of style loss $L_S$  and content loss $L_C$. 

After finding the appropriate image style, we input the appropriate style image $I_S$ and the content image $I_C$  respectively, define a loss function, this loss function style loss $L_S$ and content loss $L_C$ each account for a part, where $\alpha$ is the content loss weight, $\beta$ is the style loss weight, and the target image $I^*$ we generate needs to minimize this loss function.

The content loss $L_C$ is shown in Equation 1: 
$$ L_c =\Vert I_C-I^*\Vert_2 \eqno{(1)} $$

The style loss $L_S$ is shown in formula 2: (where $\phi_i$ represents the feature map of the $i^{th}$ layer after the image is input to the network) 
$$ L_s =\sum_{i=1} \Vert \mu(\phi_i (I^* ))-\mu(\phi_i (I_s ))\Vert_2 +\sum_{i=1}\Vert\sigma(\phi_i (I^* ))-\sigma(\phi_i (I_s ))\Vert_2  \eqno{(2)} $$

The target image $I^*$ is shown in Equation 3: 

$$I^* =\mathop{\arg\min}_{I} L_{total} (I_C,I_S,I) =\mathop{\arg\min}_{I} (\alpha L_C (I_C,I)+\beta L_S (I_S,I))\eqno{(3)}  $$
\section{Experiments}

In this section, we evaluate the effectiveness of the style transfer black-box attack method on various classification models. In particular, we aim to answer the following questions:

\begin{itemize}
	\item[$\bullet$] \textbf{RQ1} When the attack is untargeted, how is the attack performance of the proposed method compared with the state-of-the-art methods?
	\item[$\bullet$] \textbf{RQ2} How is the attack success rate of this method in the case of a targeted black box attack?
	\item[$\bullet$] \textbf{RQ3} How do parameters in the method affect the results?
	\item[$\bullet$] \textbf{RQ4} Whether the adversarial samples generated by this method have an attack effect on state-of-the-art defense models?
	\item[$\bullet$] \textbf{RQ5} How transferable are the adversarial examples generated by this method?
	\item[$\bullet$] \textbf{RQ6} Whether the adversarial samples generated by this method have changed people's judgments about the adversarial samples?
\end{itemize}

\subsection{Experimental Settings}
\subsubsection{Datasets.} The datasets used in our experiments are CIFAR-10 \cite{krizhevsky2009learning} ,CIFAR-10R and CIFAR-10NR. CIFAR-10R and CIFAR-10NR are both  extracted from CIFAR-10 by Madry \cite{carlini2017towards}. CIFAR-10R contains only robust features and CIFAR-10NR only contains non-robust features. CIFAR10 contains images for classification between classes, where the differences between different classes are large enough. For CIFAR10, we group and extract image features according to robust feature dimension /non-robust feature dimension to form datasets CIFAR-10R and CIFAR-10NR, and use these two datasets to train models that only recognize these two types of features respectively. We test our attack on these datasets  because it is easier to extract and replace non-semantic features in CIFAR-10 \cite{krizhevsky2009learning} compared to those datasets (such as MNIST\cite{xiao2017fashion}  or SVHN \cite{he2016deep}) where the non-semantic features are less different between classes.

\subsubsection{Models Used}Fourteen models were used in this experiment, which includes the  PGD adversarial training model (PGDAT) trained using PGD adversarial training, the Interpolation adversarial training model (IAT) trained using interpolation adversarial training and the other twelve models reported in Table 4-1:

\renewcommand\tabcolsep{5.0pt} % 
\renewcommand {\thetable} {4-1}
\begin{table*}
	\begin{center}
		\caption{Model list}
		\begin{tabular}{llll}
			\hline\noalign{\smallskip}
			\diagbox{Model}{Dataset}& CIFAR-10 & CIFAR-10R & CIFAR-10NR\\
			\noalign{\smallskip}
			\hline
			\noalign{\smallskip}
			Res18& RNB & RB & NRB \\
			VGG19& VGG19B & VGG19R & VGG19NR \\
			D121& D121B & D121R & D121NR \\
			GoogleNet& GNB & GNR & GNNR \\
			\hline
		\end{tabular}
	\end{center}
\end{table*}
\subsection{Untargeted Attacks}

\renewcommand {\thetable} {4-2}
\begin{table*}
	\begin{center}
		\caption{The proportion of categories judged by one thousand pictures}
		\begin{tabular}{lcllllllll}
			\hline\noalign{\smallskip}
			& Model & RNB & RB & PGDAT & IAT & NRB&VGGB&D121B&GNB\\
			\noalign{\smallskip}
			\hline
			\noalign{\smallskip}
			& Accuracy & 95.28 & 78.69 & 83.53 & 91.86 & 82.00&91.27&92.30&91.24\\
			\hline
			Attack&pgd&98.98 & 62.04& 53.93 & 55.24 & 99.90&99.53&99.48&99.76 \\
			Success & \makecell{Style\\ transfer}&95.91 & 95.11 & 78.31 & 78.36 & 92.72&89.93 &99.10&93.92\\
			\hline
		\end{tabular}
	\end{center}
\end{table*}
As shown in Table 4-2, the recognition rates of the eight models used in the first experiment are 95.28\%, 78.69\%, 83.53\%, 91.86\%, 82.00\%, 91.27\%, 92.30\%, and 91.24\%, respectively. This indicates that these models are well-trained and testable models that guarantee a certain recognition benchmark rate.

When several models are attacked by style transfer adversarial samples and PGD and FGSM adversarial samples respectively, their recognition rates dropped significantly. Among them, the less robust models ,which include RNB, NRB, VGGB, D121B and GNB, have the highest accuracy drop. Style transfer adversarial samples reduced the accuracy of these models to 4.09\%, 7.28\%, 10.07\%, 0.9\%, 6.1\%, PGD adversarial examples reduce the accuracy of these models to 1.02\% and 0.10\%, 0.47\%, 0.52\%, 0.24\%. We can see that the style transfer adversarial examples are not as good as the PGD adversarial examples in the attack effect of this type of model, but also have a good effect. The three more robust models, which include PGDAT, IAT, and RB, have less accuracy drop, and still maintain a certain accuracy. The style transfer adversarial sample reduces the accuracy of the three models to 21.69\% , 21.64\% and 4.89\% respectively. PGD adversarial examples reduce the accuracy of the three models to 46.07\%,44.76\% and 37.96\%. We can see the attack effect on this class of models, the style transfer adversarial examples far outperform the PGD adversarial examples.

It can be concluded that the untargeted attack ability of the style transfer adversarial sample on the model with strong robustness is far superior to that of the PGD adversarial sample, and the attack ability is excellent. The untargeted attack ability on the less robust model is not as good as the PGD adversarial example, but the attack ability is also strong.

\subsection{Targeted Attacks}

\renewcommand {\thetable} {4-3}
\begin{table*}
	\begin{center}
		\caption{Blackbox Target Attack Success Rate}
		\begin{tabular}{llllllllll}
			\hline\noalign{\smallskip}
			& Model & RNB & RB & PGDAT & IAT & NRB&VGGB&D121B&GNB\\
			\noalign{\smallskip}
			\hline
			\noalign{\smallskip}
			& Accuracy & 95.28 & 78.69 & 83.53 & 91.86 & 82.00&91.27&92.30&91.24\\
			\hline
			
			&Airplane&74.00	&55.62&	46.75&	59.23&	69.30&	71.12&	81.51&	69.84 \\
			&Automobile	&45.92&	43.68&	19.42&	28.32&	56.35&	55.24&	32.34&	76.07 \\
			&Bird	&58.80&	52.23&	47.95&	50.81&	48.23	&75.24&	44.35	&69.21 \\
			&Cat&	56.90	&57.21&	35.10&	32.55&	56.40&	44.82&	60.15&	8.94 \\
			Attack&Deer&	71.68&	89.63&	65.32&	73.14&	85.63&	84.21&	84.35&	96.73 \\
			&Dog&	47.14&	50.72&	42.32&	39.37&	61.52&	46.12&	45.65&	15.68 \\
			Success&Frog&	90.62&	91.63&	66.31&	74.52&	71.23&	78.76&	98.85&	86.72 \\
			&Horse&	49.53&	40.21&	25.23&	35.42&	65.21&	58.63&	52.76&	58.26 \\
			&Ship&	61.34&	87.43&	32.31&	48.23&	81.24&	66.81&	69.05&	62.48\\
			&Truck&	90.24&	82.71&	62.13&	72.35&	89.42&	92.72&	94.12&	97.58 \\
			&Total&	64.62&	65.10&	44.28&	51.40&	68.45&	67.37&	66.31&	64.15 \\
			\hline
		\end{tabular}
	\end{center}
\end{table*}

As shown in Table 4-3, the success rates of style attacks for different target categories are quite different. For styles like frog and deer, style transfer adversarial samples have a more prominent success rate of targeted attacks, which can reach more than 90\%. In the styles of dog and cat, the effect performance is not satisfactory, and the success rate of targeted attacks reach around 50\%.

Based on the experimental results, it is inferred that when the style of the target attack category is very different from other categories, such as frog and deer styles, the style transfer adversarial sample has a prominent target attack effect. However, in the case where the style of the target attack category is similar to other categories, such as cat and dog styles, the effect performance is not satisfactory.

Therefore, we believe that in the targeted attack, when the target attack type is quite different from other types of styles, our attack method is better, Besides reaching outstanding performance, our proposed method only needs to replace the style without adjusting the disturbance step by step, which benefits both efficiency and effectiveness.

\subsection{Influence of Parameters on Method Results}

\renewcommand {\thetable} {4-4}
\begin{table*}
	\begin{center}
		\caption{Adversarial samples generated under different style  parameters ($\beta$ defaults to 1)}
		\begin{tabular}{llll}
			\noalign{\smallskip}
			$\alpha=10000$& \begin{minipage}[b]{0.1\columnwidth}
				\centering
				\raisebox{-.5\height}{\includegraphics[width=\linewidth]{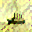}}
			\end{minipage}
			
			& $\alpha=20000$ & \begin{minipage}[b]{0.1\columnwidth}
				\centering
				\raisebox{-.5\height}{\includegraphics[width=\linewidth]{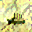}}
			\end{minipage}
			\\
			
			\noalign{\smallskip}
			$\alpha=30000$& \begin{minipage}[b]{0.1\columnwidth}
				\centering
				\raisebox{-.5\height}{\includegraphics[width=\linewidth]{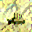}}
			\end{minipage}
			& $\alpha=40000$ & \begin{minipage}[b]{0.1\columnwidth}
				\centering
				\raisebox{-.5\height}{\includegraphics[width=\linewidth]{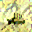}}
			\end{minipage}
			\\
			\noalign{\smallskip}
			$\alpha=50000$&\begin{minipage}[b]{0.1\columnwidth}
				\centering
				\raisebox{-.5\height}{\includegraphics[width=\linewidth]{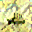}}
			\end{minipage}
			&$\alpha=60000$&\begin{minipage}[b]{0.1\columnwidth}
				\centering
				\raisebox{-.5\height}{\includegraphics[width=\linewidth]{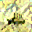}}
			\end{minipage}
			\\
			\noalign{\smallskip}
			$\alpha=70000$&\begin{minipage}[b]{0.1\columnwidth}
				\centering
				\raisebox{-.5\height}{\includegraphics[width=\linewidth]{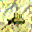}}
			\end{minipage}
			&$\alpha=80000$&	\begin{minipage}[b]{0.1\columnwidth}
				\centering
				\raisebox{-.5\height}{\includegraphics[width=\linewidth]{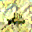}}
			\end{minipage}
			\\
			\noalign{\smallskip}
			$\alpha=90000$&\begin{minipage}[b]{0.1\columnwidth}
				\centering
				\raisebox{-.5\height}{\includegraphics[width=\linewidth]{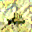}}
			\end{minipage}
			&$\alpha=100000$&\begin{minipage}[b]{0.1\columnwidth}
				\centering
				\raisebox{-.5\height}{\includegraphics[width=\linewidth]{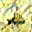}}
			\end{minipage}
			\\
			
		\end{tabular}
	\end{center}
\end{table*}
\renewcommand\tabcolsep{8.0pt} % 
\renewcommand {\thetable} {4-5}
\begin{table*}
	\begin{center}
		\caption{Adversarial sample attack success rate under different style parameters ($\beta$ defaults to 1)}
		\begin{tabular}{lll}
			\hline\noalign{\smallskip}
			& $\alpha$ & RNB\\
			\noalign{\smallskip}
			\hline
			\noalign{\smallskip}
			& 10000 & 51.00 \\
			
			&20000&68.21	 \\
			&30000	&71.42 \\
			&40000	&79.98 \\
			&50000&	81.14 \\
			Attack&60000&	82.06 \\
			Success&70000&	83.03 \\
			&80000&	86.12\\
			&90000&	85.13\\
			&100000&	84.98 \\
			\hline
		\end{tabular}
	\end{center}
\end{table*}

\subsubsection{Style weight parameters:} As shown in Table 4-4 and Table 4-5, when the style parameters continue to increase, the semantic impact on the image becomes more and more serious, and the attack success rate also increases slowly or even backwards. Choosing a good style parameter requires a good trade-off between the two.
\renewcommand\tabcolsep{3.0pt} % 
\renewcommand {\thetable} {4-6}
\begin{table*}
	\begin{center}
		\caption{Attack success rate of adversarial samples under different proportions of non-robust features and robust features}
		\begin{tabular}{llllllll}
			\hline\noalign{\smallskip}
			& proportion&\makecell{ship\\(RNB)}	&\makecell{ship\\(PGAT)}	&\makecell{truck\\(RNB)}	&\makecell{truck\\(PGAT)}	&\makecell{deer\\(RNB)}	&\makecell{deer\\(PGAT)}	\\
			\noalign{\smallskip}
			\hline
			\noalign{\smallskip}
			& 1:9&	69.05&	24.13&	66.45&	95.02&	71.68&	65.32 \\
			&3:7&	69.05&	24.13&	79.64&	97.68&	71.68&	65.32\\
			Attack&5:5&	55.25&	26.35&	79.64&	97.68&	71.68&	65.32 \\
			Success&7:3&	55.25&	26.35&	79.64&	97.68&	71.68&	65.32 \\
			&9:1&	55.25&	26.35&	79.64&	97.68&	68.23&	67.43\\
			\hline
		\end{tabular}
	\end{center}
\end{table*}

\renewcommand\tabcolsep{8.0pt} % 
\renewcommand {\thetable} {4-7}
\begin{table*}
	\begin{center}
		\caption{Comparison against defense models}
		\begin{tabular}{llll}
			\hline\noalign{\smallskip}
			& Model&	PGDAT&	IAT\\
			\noalign{\smallskip}
			\hline
			\noalign{\smallskip}
			&Accuracy&	83.53&	91.86\\
			\hline
			Attack&pgd	&53.93&	55.24	 \\
			Success&Style transfer&	78.31&	78.36 \\
			\hline
		\end{tabular}
	\end{center}
\end{table*}

\renewcommand\tabcolsep{8.0pt} % 
\renewcommand {\thetable} {4-8}
\begin{table*}
	\begin{center}
		\caption{Blackbox Non-Target Attack Success Rate}
		\begin{tabular}{llllll}
			\hline\noalign{\smallskip}
			& Model&	RNB&	VGGB&	D121B&	GNB\\
			\hline
			&Accuracy&	95.28&	91.27&	92.30&	91.24\\
			\hline
			&Style transfer(ResNet)&	95.91&	89.93&	99.10&	93.92\\
			Attack & Style transfer(VGG)&	98.11&	98.51&	99.10&	98.00\\
			Success&Style transfer(D121)&	97.92&	98.51&	99.31&	99.71 \\
			&Style transfer(GoogleNet)&	97.49&	98.40&	99.10&	99.12 \\
			\hline
		\end{tabular}
	\end{center}
\end{table*}

\renewcommand\tabcolsep{2.0pt} % 
\renewcommand {\thetable} {4-9}
\begin{table*}
	\begin{center}
		\caption{Comparison of generated adversarial examples and clean examples }
		\begin{tabular}{lllllllllll}
			\noalign{\smallskip}
			\makecell{Clean\\ Images}& \begin{minipage}[b]{0.06\columnwidth}
				\centering
				\raisebox{-.5\height}{\includegraphics[width=\linewidth]{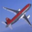}}
			\end{minipage}&
			\begin{minipage}[b]{0.06\columnwidth}
				\centering
				\raisebox{-.5\height}{\includegraphics[width=\linewidth]{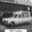}}
			\end{minipage}&
			\begin{minipage}[b]{0.06\columnwidth}
				\centering
				\raisebox{-.5\height}{\includegraphics[width=\linewidth]{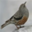}}
			\end{minipage}&
			\begin{minipage}[b]{0.06\columnwidth}
				\centering
				\raisebox{-.5\height}{\includegraphics[width=\linewidth]{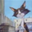}}
			\end{minipage}
			&\begin{minipage}[b]{0.06\columnwidth}
				\centering
				\raisebox{-.5\height}{\includegraphics[width=\linewidth]{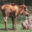}}
			\end{minipage}
			&\begin{minipage}[b]{0.06\columnwidth}
				\centering
				\raisebox{-.5\height}{\includegraphics[width=\linewidth]{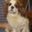}}
			\end{minipage}
			&\begin{minipage}[b]{0.06\columnwidth}
				\centering
				\raisebox{-.5\height}{\includegraphics[width=\linewidth]{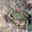}}
			\end{minipage}
			&\begin{minipage}[b]{0.06\columnwidth}
				\centering
				\raisebox{-.5\height}{\includegraphics[width=\linewidth]{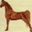}}
			\end{minipage}
			&\begin{minipage}[b]{0.06\columnwidth}
				\centering
				\raisebox{-.5\height}{\includegraphics[width=\linewidth]{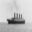}}
			\end{minipage}
			&\begin{minipage}[b]{0.06\columnwidth}
				\centering
				\raisebox{-.5\height}{\includegraphics[width=\linewidth]{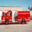}}
			\end{minipage}
			\\
			Airplane&  \begin{minipage}[b]{0.06\columnwidth}
				\centering
				\raisebox{-.5\height}{\includegraphics[width=\linewidth]{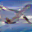}}
			\end{minipage}&
			\begin{minipage}[b]{0.06\columnwidth}
				\centering
				\raisebox{-.5\height}{\includegraphics[width=\linewidth]{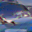}}
			\end{minipage}&
			\begin{minipage}[b]{0.06\columnwidth}
				\centering
				\raisebox{-.5\height}{\includegraphics[width=\linewidth]{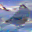}}
			\end{minipage}&
			\begin{minipage}[b]{0.06\columnwidth}
				\centering
				\raisebox{-.5\height}{\includegraphics[width=\linewidth]{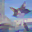}}
			\end{minipage}
			&\begin{minipage}[b]{0.06\columnwidth}
				\centering
				\raisebox{-.5\height}{\includegraphics[width=\linewidth]{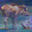}}
			\end{minipage}
			&\begin{minipage}[b]{0.06\columnwidth}
				\centering
				\raisebox{-.5\height}{\includegraphics[width=\linewidth]{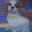}}
			\end{minipage}
			&\begin{minipage}[b]{0.06\columnwidth}
				\centering
				\raisebox{-.5\height}{\includegraphics[width=\linewidth]{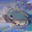}}
			\end{minipage}
			&\begin{minipage}[b]{0.06\columnwidth}
				\centering
				\raisebox{-.5\height}{\includegraphics[width=\linewidth]{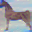}}
			\end{minipage}
			&\begin{minipage}[b]{0.06\columnwidth}
				\centering
				\raisebox{-.5\height}{\includegraphics[width=\linewidth]{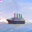}}
			\end{minipage}
			&\begin{minipage}[b]{0.06\columnwidth}
				\centering
				\raisebox{-.5\height}{\includegraphics[width=\linewidth]{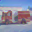}}
			\end{minipage}
			\\
			
			\noalign{\smallskip}
			Deer&   \begin{minipage}[b]{0.06\columnwidth}
				\centering
				\raisebox{-.1\height}{\includegraphics[width=\linewidth]{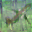}}
			\end{minipage}&
			\begin{minipage}[b]{0.06\columnwidth}
				\centering
				\raisebox{-.1\height}{\includegraphics[width=\linewidth]{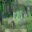}}
			\end{minipage}&
			\begin{minipage}[b]{0.06\columnwidth}
				\centering
				\raisebox{-.1\height}{\includegraphics[width=\linewidth]{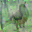}}
			\end{minipage}&
			\begin{minipage}[b]{0.06\columnwidth}
				\centering
				\raisebox{-.1\height}{\includegraphics[width=\linewidth]{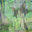}}
			\end{minipage}
			&\begin{minipage}[b]{0.06\columnwidth}
				\centering
				\raisebox{-.1\height}{\includegraphics[width=\linewidth]{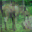}}
			\end{minipage}
			&\begin{minipage}[b]{0.06\columnwidth}
				\centering
				\raisebox{-.1\height}{\includegraphics[width=\linewidth]{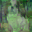}}
			\end{minipage}
			&\begin{minipage}[b]{0.06\columnwidth}
				\centering
				\raisebox{-.1\height}{\includegraphics[width=\linewidth]{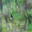}}
			\end{minipage}
			&\begin{minipage}[b]{0.06\columnwidth}
				\centering
				\raisebox{-.1\height}{\includegraphics[width=\linewidth]{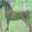}}
			\end{minipage}
			&\begin{minipage}[b]{0.06\columnwidth}
				\centering
				\raisebox{-.1\height}{\includegraphics[width=\linewidth]{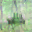}}
			\end{minipage}
			&\begin{minipage}[b]{0.06\columnwidth}
				\centering
				\raisebox{-.1\height}{\includegraphics[width=\linewidth]{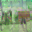}}
			\end{minipage}
			\\
			\noalign{\smallskip}
			Frog&   \begin{minipage}[b]{0.06\columnwidth}
				\centering
				\raisebox{-.5\height}{\includegraphics[width=\linewidth]{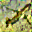}}
			\end{minipage}&
			\begin{minipage}[b]{0.06\columnwidth}
				\centering
				\raisebox{-.5\height}{\includegraphics[width=\linewidth]{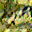}}
			\end{minipage}&
			\begin{minipage}[b]{0.06\columnwidth}
				\centering
				\raisebox{-.5\height}{\includegraphics[width=\linewidth]{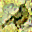}}
			\end{minipage}&
			\begin{minipage}[b]{0.06\columnwidth}
				\centering
				\raisebox{-.5\height}{\includegraphics[width=\linewidth]{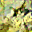}}
			\end{minipage}
			&\begin{minipage}[b]{0.06\columnwidth}
				\centering
				\raisebox{-.5\height}{\includegraphics[width=\linewidth]{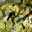}}
			\end{minipage}
			&\begin{minipage}[b]{0.06\columnwidth}
				\centering
				\raisebox{-.5\height}{\includegraphics[width=\linewidth]{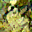}}
			\end{minipage}
			&\begin{minipage}[b]{0.06\columnwidth}
				\centering
				\raisebox{-.5\height}{\includegraphics[width=\linewidth]{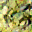}}
			\end{minipage}
			&\begin{minipage}[b]{0.06\columnwidth}
				\centering
				\raisebox{-.5\height}{\includegraphics[width=\linewidth]{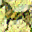}}
			\end{minipage}
			&\begin{minipage}[b]{0.06\columnwidth}
				\centering
				\raisebox{-.5\height}{\includegraphics[width=\linewidth]{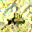}}
			\end{minipage}
			&\begin{minipage}[b]{0.06\columnwidth}
				\centering
				\raisebox{-.5\height}{\includegraphics[width=\linewidth]{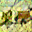}}
			\end{minipage}

			\\
			\noalign{\smallskip}
			Labels&\makecell{Airpl-\\ane}&\makecell{Autom-\\obile}	&	Bird&	cat&	deer&	dog&	frog&	horse&	ship&	truck
			\\
			
		\end{tabular}
	\end{center}
\end{table*}

\subsubsection{The proportion of non-robust features and robust features:}As shown in Table 4-6, Table 4-6 takes three categories of style attacks to conduct experiments on the proportion of non-robust features and robust features. It can be seen that in the process of increasing the proportion, the selected style has changed, so as to change the proportion of non-robust features and robust features of the style. As the proportion of robustness continues to increase, the attack effect on RNB decreases, while the attack effect on PGAT increases. Therefore, according to the experimental results, it is concluded that the larger the proportion of robust features are, the stronger the attack effect on robust models is such as PGAT, and the greater the proportion of non-robust features are, the stronger the attack effect on non-robust models is such as RNB.

\subsection{Attack Effect on Defense Models}

Currently the state-of-the-art non-authenticated defense is the adversarial training technique introduced in [11], so we evaluate the attack against the PGDAT and IAT defense techniques.

The results are shown in Table 4-7. We can see that the two defense methods of adversarial training have a certain effect on the perturbation-based basic attack, PGDAT limits the success rate to 53.93\%, IAT limits the success rate to 55.24\%. In contrast, our style-transfer adversarial examples can fool this defense more successfully, with a success rate of over 78\% under different adversarial training defense models. 
\subsection{Transferability}

An important measure of black-box adversarial attack samples is their transferability between different classifiers.

As shown in Table 4-8, the recognition rates of the four models used in the first experiment are 95.28\%, 91.27\%, 92.30\%, and 91.24\%, respectively. This indicates that these models are well-trained and testable models, which guarantee a certain recognition benchmark rate. The experiment generates corresponding adversarial samples for all four models, and uses these generated adversarial samples to attack these models. We can see that the attack success rate of the adversarial samples generated by the ResNet model for RNB, VGGB, D121B, and GNB is basically over 90\%. The VGG model, D121 model and GoogleNet model also have more than 90\% attack success rate for RNB, VGGB, D121B and GNB. It can be seen that the adversarial samples based on style transfer have transferability and the transferability is strong.
\subsection{Adversarial Example Visual Effects}

Table 4-9 shows examples of using adversarial transformations to generate corresponding adversarial examples on the CIFAR10 dataset. The first row in the figure represents the original clean samples with corresponding visual labels, and the next few rows represent the adversarial samples generated by the style transfer method for targeted attacks on different categories. The size of the image in the CIFAR10 dataset is 32 × 32, and the visualization resolution is low. From the figure, it can be found that the style transfer adversarial transformation modifies the background information relatively more. However, the shape characteristics of the objects in the picture are still well preserved, that is, the semantic information of the picture is preserved, and the human eye can visually recognize the target object accurately.

In analogy to the human visual cognitive process, humans pay more attention to the correlation between image semantics during recognition, which is helpful for understanding and recognizing objects. While convolutional networks give more weight to style recognition in the recognition process than humans, which is easier to be deceived and misled by the replaced style features.

\section{Conclusions}

In this paper, we directly start with non-semantic features for feature replacement, and propose a new method to generate adversarial examples through image style feature replacement. We pick the most suitable style feature from each target class only by judging the style feature confidence of the RB model and the NRB model. In other words, the style features are selected from two dimensions, the robust feature dimension and the non-robust feature dimension. This style feature replacement attack is common to each class. Since it's only needed to pick out the appropriate style features without having to find the perturbations corresponding to these style features, The process of generating style transfer adversarial examples is efficient. And this method is a black-box model, that is, it does not need to know the parameters inside the attacked model, or even know what kind of network structure the attacked model is,nor need the confidence output of the input image like ZOO\cite{chen2017zoo}. It is a pure black-box model. In addition, in the case of targeted attack, this model has better effect on the target class whose style is quite different from other classes, while the effect of other classes is average, so it is suitable for image classification with large style differences between classes. In the future , finding important non-semantic features for replacement from different latitudes will be another important and promising direction.

\appendix

\bibliographystyle{splncs04}
\bibliography{egbib}

\end{document}